\documentclass[letterpaper, 10 pt, conference]{ieeeconf}  

\IEEEoverridecommandlockouts                              

\usepackage{graphics} 
\usepackage{epsfig} 
\usepackage{times} 
\usepackage{amsmath} 
\usepackage{amssymb}  
\usepackage{mathrsfs}
\usepackage{algorithm}
\usepackage{algorithmicx}
\usepackage{algpseudocode}
\usepackage{cite}
\usepackage{booktabs}
\usepackage{epstopdf}
\usepackage{subfigure,color,balance}
\usepackage{verbatim}
\usepackage{cases}
\usepackage{enumerate}
\usepackage{balance}
\usepackage{threeparttable}
\usepackage{stfloats}

\usepackage[colorlinks=true,      
linkcolor=black,      
citecolor=black,      
filecolor=black,      
urlcolor=blue]{hyperref}

\def\0{{\bf 0}}
\def\1{{\bf 1}}

\title{\LARGE \bf
TeraSim-World: Worldwide Safety-Critical Data Synthesis for End-to-End  Autonomous Driving}

\author{Jiawei Wang$^{1\dagger}$, Haowei Sun$^{2\dagger}$, Xintao Yan$^{3}$, Shuo Feng$^{4}$, Jun Gao$^{1,5}$, Henry X. Liu$^{1*}$
\thanks{This work was partially funded by the United States National Science
Foundation through the Mcity 2.0 Project (CMMI \#2223517). $^\dagger$Equal contribution. $^*$Corresponding author. (email: henryliu@umich.edu)}
\thanks{$^{1}$University of Michigan. $^{2}$SaferDrive AI. $^{3}$The University of Hong Kong. $^{4}$Tsinghua University. $^{5}$NVIDIA. }
}

\begin{document}

\maketitle
\thispagestyle{empty}
\pagestyle{empty}

\begin{abstract}


Safe and scalable deployment of end-to-end (E2E) autonomous driving requires extensive and diverse data, particularly safety-critical events. Existing data are mostly generated from simulators with a significant sim-to-real gap or collected from on-road testing that is costly and unsafe. This paper presents TeraSim-World, an automated pipeline that synthesizes realistic and geographically diverse safety-critical data for E2E autonomous driving at anywhere in the world. Starting from an arbitrary location, TeraSim-World retrieves real-world maps and traffic demand from geospatial data sources. Then, it simulates agent behaviors from naturalistic driving datasets, and orchestrates diverse adversities to create corner cases. Informed by street views of the same location, it achieves photorealistic, geographically grounded sensor rendering via the frontier video generation model  Cosmos-Drive. By bridging agent and sensor simulations, TeraSim-World provides a scalable and critical~data synthesis framework for training and evaluation of E2E autonomous driving systems. Codes and videos are available  \href{https://wjiawei.com/terasim-world-web/}{here}.

\end{abstract}

\section{Introduction}

Despite recent advances of end-to-end (E2E) autonomous driving, two fundamental challenges remain for real-world deployment of such autonomous vehicle (AV) systems: scalability and safety~\cite{liu2025autonomous,chen2024end}. Scalability refers to the capability to generalize reliably across diverse and previously unseen operational design domains (ODDs), while safety ensures robust operation under not only normal traffic environments but also rare yet safety-critical scenarios. Existing datasets such as Waymo Open Dataset~\cite{sun2020scalability} and nuScenes~\cite{caesar2020nuscenes} are geographically restricted, scenario-constrained, and highly imbalanced toward normal driving rather than corner cases. As a result, current AV development pipelines often demand new data collection and annotation when deploying in new cities or countries, and still rely heavily on extensive on-road testing with the aim of encountering diverse conditions and exposing safety-critical events. Such real-world data collection and testing, however, is prohibitively costly, time inefficient, and inherently unsafe. 


Thanks to recent advances in generative AI, data synthesis has emerged as a compelling and cost-effective alternative to expensive and risky on-road data collection and testing. It enables scalable generation of diverse, and even safety-critical, driving events that can be used to train and evaluate E2E autonomous driving systems. Along this direction, agent simulation and sensor simulation constitute the two main components. Agent simulation focuses on modeling the trajectories of traffic participants. Early approaches focus on rule-based models~\cite{treiber2013traffic}, with SUMO~\cite{krajzewicz2012recent} serving as a representative platform. More recently, learning behavior models directly from naturalistic driving datasets has enabled more realistic trajectory-level simulation (see, \emph{e.g.}, auto-regressive frameworks\cite{wu2024smart,yan2023learning,zhang2025closed} and diffusion-based methods\cite{huang2024versatile,zhong2023guided,jiang2024scenediffuser}), and further allowed for controllable synthesis of adversarial behaviors in safety-critical scenarios~\cite{xu2025diffscene,chang2024safe,wang2025rade}.

Despite this progress, agent simulation typically produces trajectories alone. Without rendering them into sensor-level inputs such as camera or lidar, those synthesized data cannot be directly compatible with E2E driving stacks. 
Video generation addresses this limitation by producing photorealistic visual data. Recent advances in video diffusion models~\cite{ho2022video} have enabled temporally coherent multi-view video synthesis~\cite{wang2024drivedreamer,gao2023magicdrive,zhao2025drivedreamer}. However, most existing approaches predict videos only from past frames \cite{voleti2022mcvd,hoppe2022diffusion} or condition on limited motion information such as ego actions \cite{hu2023gaia,wang2024driving} or initial traffic layouts \cite{wen2024panacea,yang2024drivearena}. Without explicit integration of full agent trajectories, they struggle to ensure physically plausible outputs, especially in safety-critical scenarios. In addition, they typically produce only short, low-frame-rate clips, and rely on small datasets such as nuScenes, which limits their scalability to new cities and diverse conditions.

\begin{figure*}[htbp!]
	\centering
	{\includegraphics[width=0.99\textwidth]{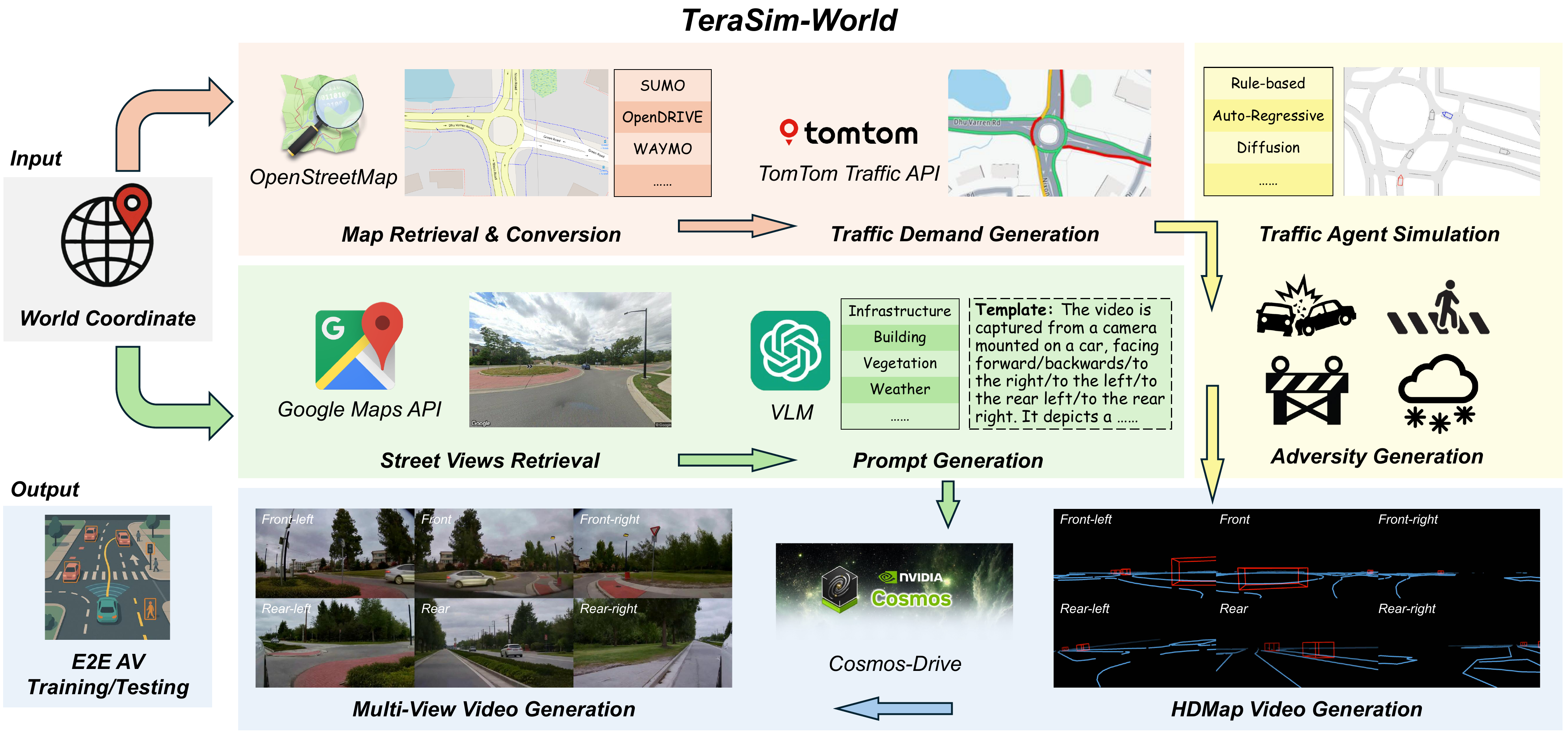}}
	\vspace{-2mm}
	\caption{TeraSim-World Pipeline. }
    \vspace{-6mm}
	\label{Fig:Framework}
\end{figure*}

To bridge controllable agent simulation with realistic sensor simulation, this paper presents TeraSim-World, an automatic data synthesis pipeline for E2E autonomous driving at anywhere in the world. The framework is shown in Fig.~\ref{Fig:Framework}. Starting from an arbitrary global coordinate, TeraSim-World automatically retrieves the real-world map and traffic demand, and collects local street-view imagery to provide environmental context. For agent simulation, it leverages the generative behavior modeling framework of TeraSim~\cite{sun2025terasim}, supporting both naturalistic and adversarial behaviors. Unlike most agent simulation platforms that rely on virtual simulators like CARLA~\cite{dosovitskiy2017carla} for sensor rendering, TeraSim-World integrates trajectory generation with multi-view video synthesis  using the frontier video generation model Cosmos-Drive~\cite{ren2025cosmos}. Fine-tuned from NVIDIA’s Cosmos world foundation models~\cite{agarwal2025cosmos}, Cosmos-Drive generates photorealistic multi-view driving videos with high resolution and spatial-temporal consistency, conditioned on structured inputs like HDMap videos. By unifying these components, TeraSim-World enables scalable and geographically grounded synthesis of realistic and safety-critical driving data. The key contributions of this work are  as follows:

\newpage 

\begin{itemize}
\item \textbf{Bridging Agent and Sensor Simulations:} TeraSim-World supports trajectory generation of traffic agents (including vehicles, cyclists, and pedestrians) by extending the behavior modeling framework of TeraSim. These trajectories, including normal and adversarial interactions, are then rendered into multi-view HDMap videos and transformed into photorealistic driving videos via Cosmos-Drive. By bridging trajectory generation with sensor rendering, our pipeline ensures that visual outputs remain aligned with underlying agent behaviors. This capability enables \textit{controllable and plausible synthesis of safety-critical E2E driving data}.

\item \textbf{Automated and Geographically Grounded Pipeline:}
TeraSim-World is a fully automated data synthesis pipeline, requiring no manual configuration for generating normal or adversarial events. Due to modularized design, each module can be replaced  with alternative or more advanced methods. Particularly, it is geographically grounded, with every component informed by real-world data: maps and traffic demand are obtained from geospatial data sources, and video generation is inspired by local street view images using vision-language models (VLMs). These properties enable \textit{scalable and realistic event generation at anywhere in the world}.


\end{itemize}

\section{Related Work}

\subsection{Traffic Agent Simulation}

Traffic agent simulation aims to generate agent-level trajectories that realistically capture interactions among traffic participants. Early simulators rely on rule-based models, such as car-following and lane-changing approaches in SUMO~\cite{krajzewicz2012recent}. While effective for reproducing aggregate flow dynamics, these models lack the capability of reproducing the variability and stochasticity of real human driving. To address this gap, recent work has shifted towards data-driven behavior modeling, where models are learned directly from naturalistic driving datasets. Imitation learning has emerged as a key technique, and particularly, 
foundation-model-inspired approaches have recently shown remarkable progress. GPT-style auto-regressive models~\cite{wu2024smart,zhang2025closed,yan2023learning} leverage the capability of transformers in prediction of tokenized future trajectories to model multi-agent dynamics. Diffusion-based approaches~\cite{zhong2023guided,jiang2024scenediffuser,huang2024versatile} have demonstrated strong controllability and scene consistency. Open challenges such as the Waymo Open Sim Agent Challenge~\cite{montali2023waymo} have further accelerated progress by providing standardized benchmarks for algorithm evaluation and iteration. 

Beyond normal driving behaviors, agent simulation must also capture  adversarial interactions that rarely occur in the real world but are critical for AV testing. One popular direction is to leverage diffusion-based models for controllable safety-critical behavior generation. For instance, DiffScene~\cite{xu2025diffscene} and Safe-Sim~\cite{chang2024safe} employ guided diffusion sampling to perturb normal behaviors into high-risk ones, and RADE~\cite{wang2025rade} introduces risk-conditioned diffusion to model multi-agent environment under desired risk levels. Most prior works address either normal or adversarial driving behaviors in isolation, and very recently, TeraSim~\cite{sun2025terasim} adopts the Naturalistic and Adversarial Driving Environment (NADE), a unified framework that augments naturalistic flows with adversarial behaviors regulated by dense reinforcement learning~\cite{feng2023dense}. Based on an adversity orchestrator for global control over timing and frequency, this framework ensures adversarial exposure with statistical realism, providing a scalable and balanced approach to stress-testing AVs under both normal and safety-critical conditions.

\subsection{Video Generation Models}

Sensor simulation is essential for data synthesis to train and evaluate perception modules and E2E autonomous driving stacks. Recent progress in video diffusion models has enabled driving video generation with improved realism. Early approaches primarily focus on predicting future frames from historical observations~\cite{voleti2022mcvd,hoppe2022diffusion}, while more recent models have begun to incorporate motion information. For example, GAIA-1~\cite{hu2023gaia} and Drive-WM~\cite{wang2024driving} generate future frames conditioned on ego-vehicle actions, whereas DriveDreamer~\cite{wang2024drivedreamer}, MagicDrive~\cite{gao2023magicdrive}, and Panacea~\cite{wen2024panacea} condition on initial traffic layouts or bird’s-eye-view (BEV) representations. 
Despite these advances, most existing models generate only short video clips at low frame rates, which fall short of the requirements of AV stacks. Since their conditioning usually covers only an initial layout or a short segment, they often fail to maintain alignment with full driving trajectories, which is critical for plausible video generation, particularly in corner cases. In addition, most existing models are trained on small-scale datasets such as nuScenes, which restricts their geographic scalability and cannot generalize to new environments or unseen scenarios.

Very recently, Cosmos-Drive~\cite{ren2025cosmos} has emerged as a new frontier for driving video generation. Fine-tuned from NVIDIA’s Cosmos world foundation models~\cite{agarwal2025cosmos} on large-scale driving data, Cosmos-Drive enables high-resolution, high-frame-rate multi-view video synthesis, conditioned on structured inputs such as HDMap videos. The model achieves spatially and temporally coherent outputs across all camera perspectives, substantially surpassing prior methods in both scalability and fidelity. This capability opens up new opportunities for geographically realistic and scalable visual data synthesis at diverse driving scenarios.

\section{Framework}

TeraSim-World is a modular and fully automated pipeline designed to generate realistic safety-critical data for E2E autonomous driving. The  framework is illustrated in Figure~\ref{Fig:Framework}. 

Given an arbitrary global coordinate, TeraSim-World first retrieves the real-world road map. This raw map is then converted into simulation-ready formats, enabling integration with different agent behavior modeling backends. The traffic demand generation module determines background traffic patterns from live traffic sources.

Meanwhile, the prompt generation module extracts semantic scene descriptions from street view imagery using VLM. These prompts capture static environmental features and ensure that the generated videos reflect the geographic context of the chosen location.

The agent simulation module then generates the trajectories for traffic participants. This module is extended from  TeraSim~\cite{sun2025terasim}, which simulates realistic multi-agent trajectories, and orchestrates adversarial interactions to inject safety-critical events. The output is a set of agent trajectories in the chosen map layout.

To render sensor-level outputs, the agent trajectories and map layouts are first rendered into multi-view HDMap videos. These serve as structured control inputs for Cosmos-Drive~\cite{ren2025cosmos}. Then, Cosmos-Drive synthesizes temporally coherent and geographically grounded multi-view videos. The resulting data can be used for training, testing, and benchmarking E2E autonomous driving systems.

\section{Methodology}

In this section, we introduce the implementation of each module in TeraSim-World.

\subsection{Real-World Map and Traffic}

We begin by introducing the key components of TeraSim-World that enable realistic map retrieval and traffic demand generation based on real-world data.

\vspace{0.2em}
\noindent \textbf{Map Retrieval:}
To support geographically grounded simulation, TeraSim-World supports automatic retrieval of real-world maps from OpenStreetMap (OSM). Users can choose from three flexible retrieval modes:

\begin{itemize}
\item \textit{Geographic Coordinates:} Specify latitude and longitude with a defined radius to extract precise local areas. This is ideal for replicating crash reports, news-related events, or known regions of interest.

\item \textit{City and Road Type:} Input a city name (\textit{e.g.}, Ann Arbor, MI, or Chicago, IL) and filter by road types (\textit{e.g.}, highways, roundabouts, intersections). This enables retrieval of targeted road topology in specific regions.

\item \textit{Travel Route:} Define an origin-destination pair to generate a point-to-point driving route. This supports trip-level simulations and evaluations.
\end{itemize}

\vspace{0.2em}
\noindent \textbf{Map Conversion:} After retrieving the OSM data, TeraSim-World provides automatic conversion to multiple map formats to support a wide range of agent simulation backends:

\begin{itemize}
\item \textit{SUMO:} To support simulation in standard traffic simulators of SUMO~\cite{krajzewicz2012recent} with rule-based behavior modeling and route planning for various traffic participants.
\item \textit{OpenDRIVE:} To integrate with physics-based simulators like CARLA~\cite{dosovitskiy2017carla} and support HD map elements such as lane geometry and road semantics.
\item \textit{Waymo:} To support data-driven behavior modeling methods built upon the Waymo Open Dataset format~\cite{montali2023waymo}, and related datasets and tools.
\end{itemize}

This  automated map retrieval and conversion pipeline minimizes manual preprocessing and ensures seamless interoperability across different simulators.


\vspace{0.2em}
\noindent \textbf{Traffic Demand Generation:}
To generate traffic demand in the road network, TeraSim-World supports two modes:
\begin{itemize}
    \item \textit{Manual Specification:} In the basic approach of manual specification, users can define desired levels of traffic demand for different types of road users, including vehicles, pedestrians, and cyclists. This is standardized in simulators like SUMO to generate traffic flows.
    \item \textit{Real-World Data-Informed Generation:} To further capture real-world traffic patterns, TeraSim-World integrates live traffic data from \textit{TomTom Traffic API}. For each road segment in the network, the API provides both the current traffic speed and the nominal free-flow speed. By comparing these values, one can estimate traffic density based on  macroscopic traffic flow models, such as Greenshields Model~\cite{treiber2013traffic}. This information allows for automatic generation of traffic demand. 
\end{itemize}

Compared with manual specification, the data-informed approach allows for geographically grounded traffic patterns, capturing real-world effects such as rush-hour congestion and localized bottlenecks.

\subsection{Agent Simulation}

Given the road map and traffic demand, TeraSim-World generates the trajectories of individual traffic agents through behavior simulation, which consists of normal behavior modeling and safety-critical adversity generation.

\vspace{0.2em}
\noindent \textbf{Behavior Modeling:} 
For high fidelity simulation, it is critical to model multi-agent driving behaviors with statistical realism, which is also known as building the \textit{Naturalistic Driving Environment (NDE)}~\cite{yan2023learning}. Depending on the selected map format, multiple behavior modeling options are supported:

\begin{itemize}
\item For SUMO or OpenDRIVE maps, TeraSim-World enables rule-based simulations using embedded behavior models in SUMO and CARLA.
\item For Waymo-format maps, data-driven approaches from the Waymo Open Sim Agents Challenge (WOSAC)~\cite{montali2023waymo} can be directly integrated.
\end{itemize}

Compared with rule-based methods, data-driven models from large-scale naturalistic driving datasets provide higher-fidelity and more diverse agent behaviors. The current pipeline has supported open-source methods like SMART~\cite{wu2024smart} and VBD~\cite{huang2024versatile}, both of which achieved leading performance in WOSAC 2024. With native compatibility for the Waymo format, TeraSim-World can seamlessly integrate any future agent simulation models developed within this ecosystem, ensuring extensibility as the field advances.

\vspace{0.2em}
\noindent \textbf{Adversity Generation:}
One core capability of TeraSim-World is to synthesize safety-critical scenarios that challenge the limits of AV stacks. To achieve this, TeraSim-World extends the adversity generation mechanism from TeraSim~\cite{sun2025terasim}, which consists of two main components:
\begin{itemize}
    \item \textit{Adversity Orchestrator:} Control the injection of adversarial events, including static adversities like road construction zones and weather conditions, and dynamic adversities involving background traffic participants.
    \item \textit{Naturalistic and Adversarial Driving Environment (NADE):} Regulate the frequency of adversities using real-world crash statistics and dense reinforcement learning~\cite{feng2023dense}. This ensures statistically representative exposure to long-tail safety-critical events without compromising overall realism.
\end{itemize}

We formalize an adversity-injection module with three elements: trigger conditions, activated behaviors, and activation probabilities. This supports the targeted and automatic injection of plausible risk factors into the environment, such as cut-in, failure to yield, and red-light running involving both cars and vulnerable road users like pedestrians and cyclists. Interested readers are referred to TeraSim~\cite{sun2025terasim} for implementation in the Mcity test facility environment.

\subsection{Sensor Simulation}

To enable sensor simulations for onboard cameras, TeraSim-World synthesizes high-fidelity driving videos aligned with structured map and trajectory information. This capability is powered by Cosmos-Drive~\cite{ren2025cosmos}, which is a series of models post-trained from Cosmos-1 World Foundation Models (WFMs)~\cite{agarwal2025cosmos} for autonomous driving scenarios.

\vspace{0.2em}
\noindent \textbf{Cosmos-Drive:}
Cosmos-1 is a suite of generalist WFMs for Physical AI, developed by NVIDIA.  Within this ecosystem, Cosmos-Drive comprises a series of driving-specific video diffusion models, post-trained on large-scale driving datasets to support controllable, high-fidelity video generation. Specifically, it leverages two key datasets: 1) Real Driving Scene (RDS), which comprises approximately 20,000 hours of six-view driving videos; 2) Real Driving Scene HQ (RDS-HQ), containing 750 hours of driving videos with corresponding HDMap and 3D cuboid annotations. 

In TeraSim-World, we integrate two specialized models: \textit{Cosmos-Transfer1-7B-Sample-AV} for front view video generation and \textit{Cosmos-7B-Single2Multiview-Sample-AV} for surrounding view synthesis, both conditioned on text prompts and HDMap videos to produce photorealistic and semantically grounded driving videos.

\begin{table}[t!]
\footnotesize
	\begin{center}
		\caption{Camera Parameters for Google Street View (Unit: deg)} 
		\vspace{-1mm}
		\label{tab:streetview_params}
		\begin{threeparttable}
		\setlength{\tabcolsep}{3.5mm}{
		\begin{tabular}{ccccccc}
		\toprule
			& F & FL & FR & R & RL & RR \\
\midrule
Heading & 0   & -66  & 66  & 180  & -152 & 152 \\
FOV          & 120 & 120  & 120 & 30   & 70   & 70  \\
			\bottomrule
		\end{tabular}}
		\begin{tablenotes}
		\footnotesize
		\item F, FL, FR, R, RL, and RR represents front, front-left, front-right, rear, rear-left, and rear-right, respectively.
		\end{tablenotes}
		\end{threeparttable}
	\end{center}
	\vspace{-4mm}
\end{table}

\vspace{0.2em}
\noindent \textbf{VLM-based Prompt Generation:}
To generate geographically grounded videos, TeraSim-World designs a prompt generation pipeline based on street view images and VLMs. Given a global coordinate, the \textit{Google Street View Static API} is queried to retrieve images using configurable camera parameters such as heading and field of view. The parameters for the six surrounding views are listed in Table~\ref{tab:streetview_params}, chosen to closely match the real multi-camera setup in the RDS-HQ dataset. This ensures that each retrieved street-view image corresponds to one onboard camera perspective.

The retrieved street views are then processed by VLMs to extract natural language descriptions of the static environment. We employ \textit{GPT-4o} in the current pipeline, and focus on scene context such as infrastructure, buildings, and vegetation, while ignoring traffic participants in the original street views to avoid biasing the prompt. The resulting text descriptions are passed to Cosmos-Drive to condition the visual style and semantic content of the generated videos. This automated pipeline enables video synthesis that is both geographically grounded and visually consistent with the real-world environment, supporting realistic and location-aware video generation. In addition, weather-related adversities  can be embedded into the prompts.

\begin{figure}[t]
	\vspace{1mm}
	\centering
	{\includegraphics[width=0.38\textwidth]{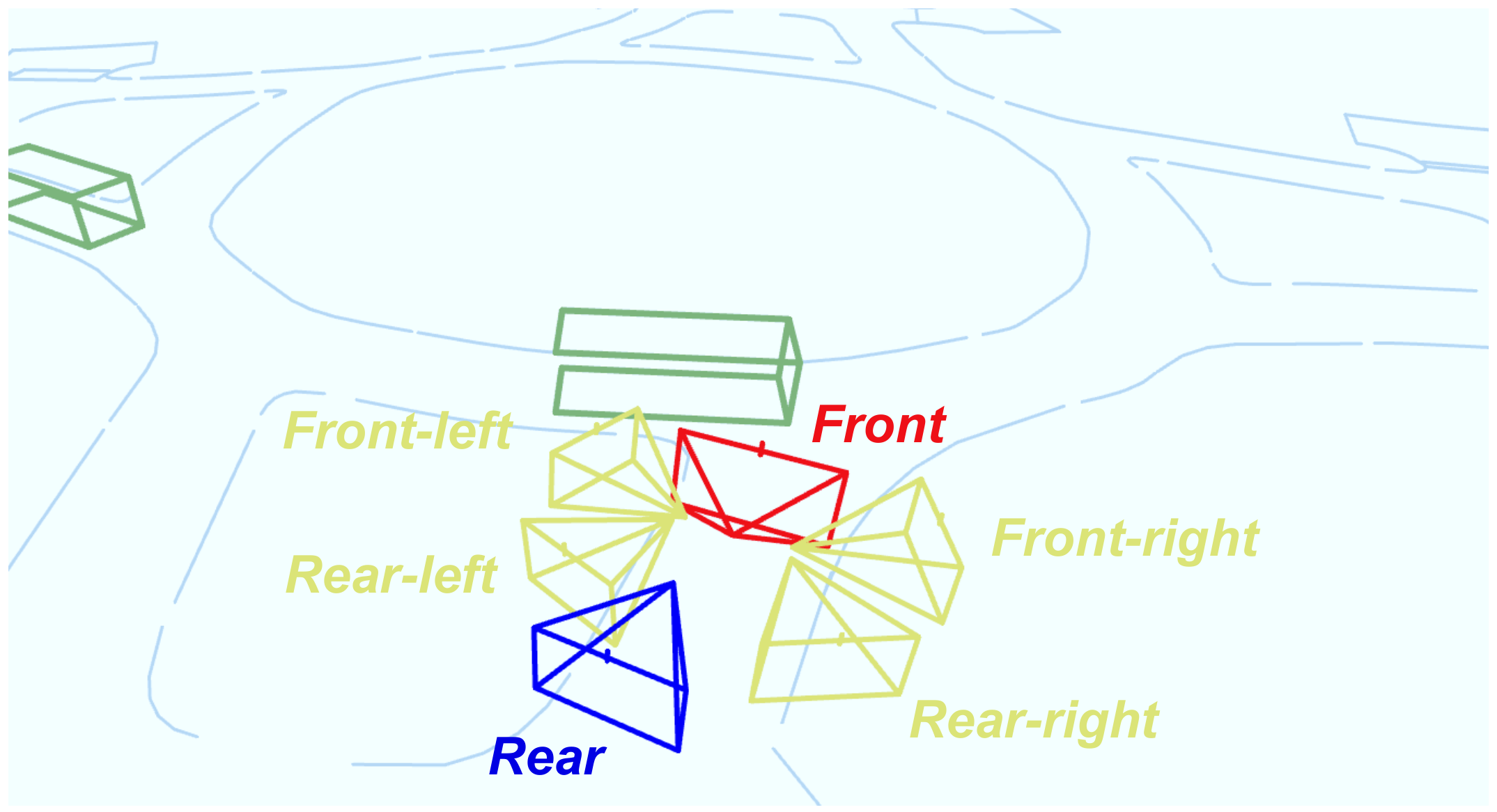}}
	\vspace{-2mm}
	\caption{Illustration of view projection for HDMap video rendering.}
	\label{Fig:hdmap_render}
    \vspace{-5mm}
\end{figure}

\vspace{0.2em}
\noindent \textbf{HDMap Video Rendering:}
Given the road map and agent trajectories generated in earlier modules, we next render HDMap videos to serve as conditional inputs for the Cosmos models. Specifically, we render the road network elements (including lane lines, road boundaries, and crosswalks) in a 3D space, and represent traffic participants as 3D cuboids. The 3D scene is then projected into 2D video frames using the intrinsic and extrinsic parameters of the six surrounding cameras from the RDS-HQ dataset. An illustration of the camera setup is provided in Fig.~\ref{Fig:hdmap_render}, which shows the position and coverage of each camera. The resulting HDMap videos encode structured spatial-temporal information and are formatted to match the input specification of Cosmos.

\vspace{0.2em}
\noindent \textbf{Frontal-View Generation:}
We then utilize the conditional video generation model of \textit{Cosmos-Transfer1-7B-Sample-AV}  to generate frontal-view videos using the HDMap video and text prompts. Cosmos-Transfer1 builds upon the Cosmos-Predict1~\cite{agarwal2025cosmos} world model using a diffusion transformer (DiT) based architecture. The base model is extended with ControlNet~\cite{zhang2023adding} branches that handle conditional inputs. For HDMap conditioning, the rendered HDMap video is tokenized and passed through a dedicated control branch composed of transformer blocks. These control activations are integrated into the main generation pipeline using zero-initialized linear layers, ensuring the spatial structure of the control input guides the generation process. 

\vspace{0.2em}
\noindent \textbf{Multi-View Generation:}
To extend from frontal-view generation to surrounding-view synthesis, we adopt \textit{Cosmos-7B-Single2Multiview-Sample-AV}. This model introduces a view-extension mechanism by leveraging clean video tokens for conditioning inputs and appending per-view indicators to distinguish between input and target views. Tokens from all views are concatenated and processed jointly using DiT blocks, allowing for cross-view attention and temporal coherence. Each view is paired with its own text prompt, which is independently embedded to guide style and semantics.  For layout preservation across multiple views, a multi-view ControlNet is also trained using HDMap-based scene representations. This architecture allows for multi-view video generation with spatial-temporal consistency.

\begin{figure}[t]
	\vspace{1mm}
	\centering
	{\includegraphics[width=0.46\textwidth]{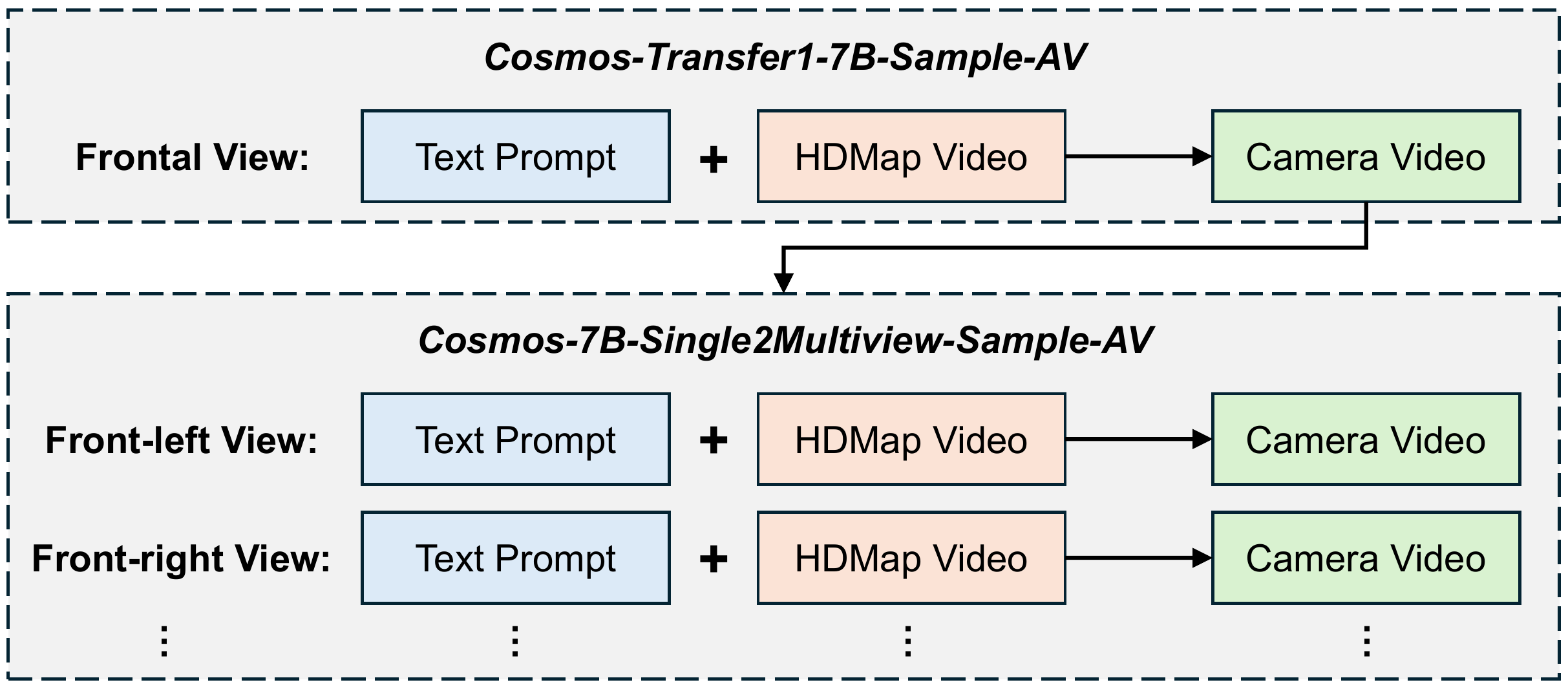}}
	\vspace{-2mm}
	\caption{Multi-View Video Generation Pipeline.}
	\label{Fig:single_to_multi}
    \vspace{-3mm}
\end{figure}

The complete pipeline for multi-view video generation is illustrated in Fig.~\ref{Fig:single_to_multi}. The front-view video is first generated using the text prompt and HDMap video via \textit{Cosmos-Transfer1-7B-Sample-AV}. This generated front-view video is then used as an additional input, together with HDMap videos and prompts of other views, to condition \textit{Cosmos-7B-Single2Multiview-Sample-AV}, which produces  videos across all surrounding camera views. The front-view video is at $24$ FPS with $120$ frames in total and a resolution of $704 \times 1280$, while the multi-view output contains six synchronized streams at $24$ FPS with $57$ frames per view ($342$ frames in total) and a per-view resolution of $576 \times 1024$.


\section{Results}
\label{Sec:5}

This section presents examples of generated safety-critical data at different locations. For clarity, TeraSim-World can automatically synthesize a variety of safety-critical events without manual intervention, and we present only one representative case per location in the following.

\begin{figure*}[t]
	\vspace{1mm}
	\centering
    \includegraphics[width=0.98\textwidth]{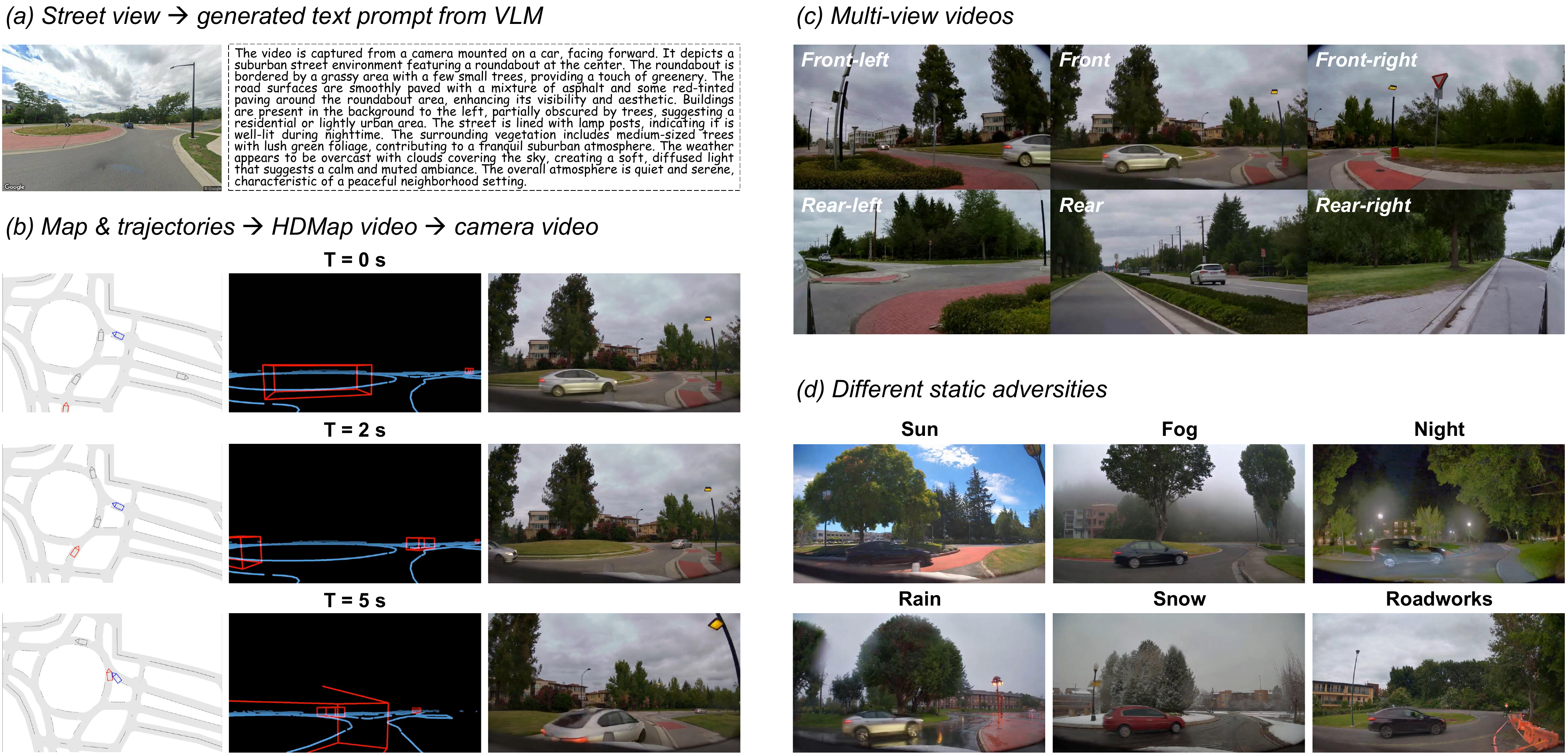}
    \vspace{-2mm}
    \caption{Case study at a roundabout in Ann Arbor, U.S. (42.31674\textdegree N, 83.70770\textdegree W). In (a) and (b) we show the frontal view as an example; for all six views, the pipeline provides the street-view image, generated prompts, rendered HDMap videos, and final photorealistic videos.}
    \vspace{-3mm}
	\label{Fig:case_study}
\end{figure*}

\begin{figure}[!t]
	\centering
     \includegraphics[width=0.48\textwidth]{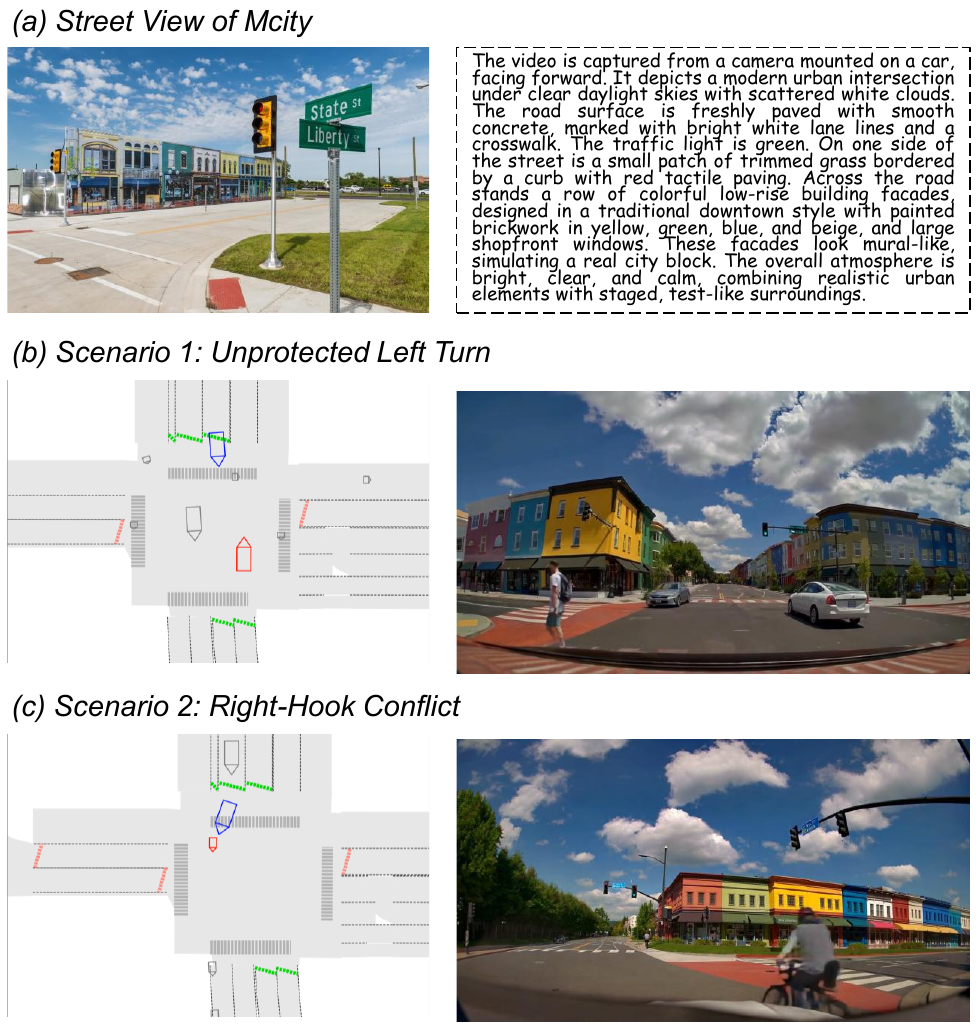}
     \vspace{-6mm}
	\caption{Safety-critical events at Mcity test facility.}
	\label{Fig:mcity_case}
    \vspace{-4mm}
\end{figure}

\subsection{Case Study: Roundabout in Ann Arbor, U.S.}

We first show a case study at a roundabout in Ann Arbor, U.S. Fig.~\ref{Fig:case_study}(a) shows the real-world street view image retrieved from Google Maps, as well as the text prompt generated by the VLM model, providing the environmental context for video synthesis. Here we show the front-view example, and the street-view images and prompts are obtained for all six perspectives.

Fig.~\ref{Fig:case_study}(b) provides the snapshot for the simulated trajectories, and the rendered HDMap videos and the generated videos for the front-view camera at different time steps. In this scenario, the ego vehicle waits at the entrance, yields to two vehicles, and then enters despite another vehicle already circulating, creating a high-risk fail-to-yield maneuver. The resulting video demonstrates that the motions of surrounding vehicles closely follow the simulated trajectories, indicating strong alignment between agent simulation and sensor simulation. Moreover, the visual style and environment are consistent with the real-world street view, effectively resembling the actual roundabout at that location.  Fig.~\ref{Fig:case_study}(c) further presents multi-view synthesis results, which remain spatially and temporally consistent across six views and align well with the real-world context of the roundabout.

We proceed to explore static adversities by applying variations in environmental conditions. As shown in Fig.~\ref{Fig:case_study}(d), we demonstrate synthesized videos under different weather conditions, as well as roadworks. These examples show the flexibility of TeraSim-World in generating diverse static adversarial events of the same base scenario.

\subsection{From Test Facility to Real World}

In the previous work TeraSim~\cite{sun2025terasim}, safety-critical cases were generated in Mcity and rendered in CARLA. As a typical test facility, Mcity provides a controlled environment, but its miniature setup and limited fidelity cannot fully reflect the real-world complexity. Similarly, CARLA rendering, though useful for closed-loop testing, suffers from a noticeable sim-to-real gap in terms of sensor-level fidelity.

With TeraSim-World, we extend these cases from Mcity into real-world style videos. Using a single picture from Mcity as the “street view” reference (Fig.~\ref{Fig:mcity_case}(a)), our pipeline retrieves environmental context and produces photorealistic camera videos for similar corner cases. The first case is an unprotected left turn (Fig.~\ref{Fig:mcity_case}(b)): the ego vehicle attempts to turn left but must stop, as a pedestrian is crossing and another vehicle is approaching from the opposite direction. The second case is a right-hook conflict (Fig.~\ref{Fig:mcity_case}(c)): as the ego vehicle turns right, a cyclist emerges from the blind spot on the right-hand side, creating a dangerous interaction. These scenarios, originally rendered in a simplified Mcity-CARLA setup, can now be visualized in geographically grounded videos, demonstrating TeraSim-World’s ability to bridge controlled test facilities with realistic, safety-critical agent and sensor simulation environments.

\begin{figure*}[!hbtp]
	\vspace{1mm}
	\centering
    \includegraphics[height=0.96\textheight]{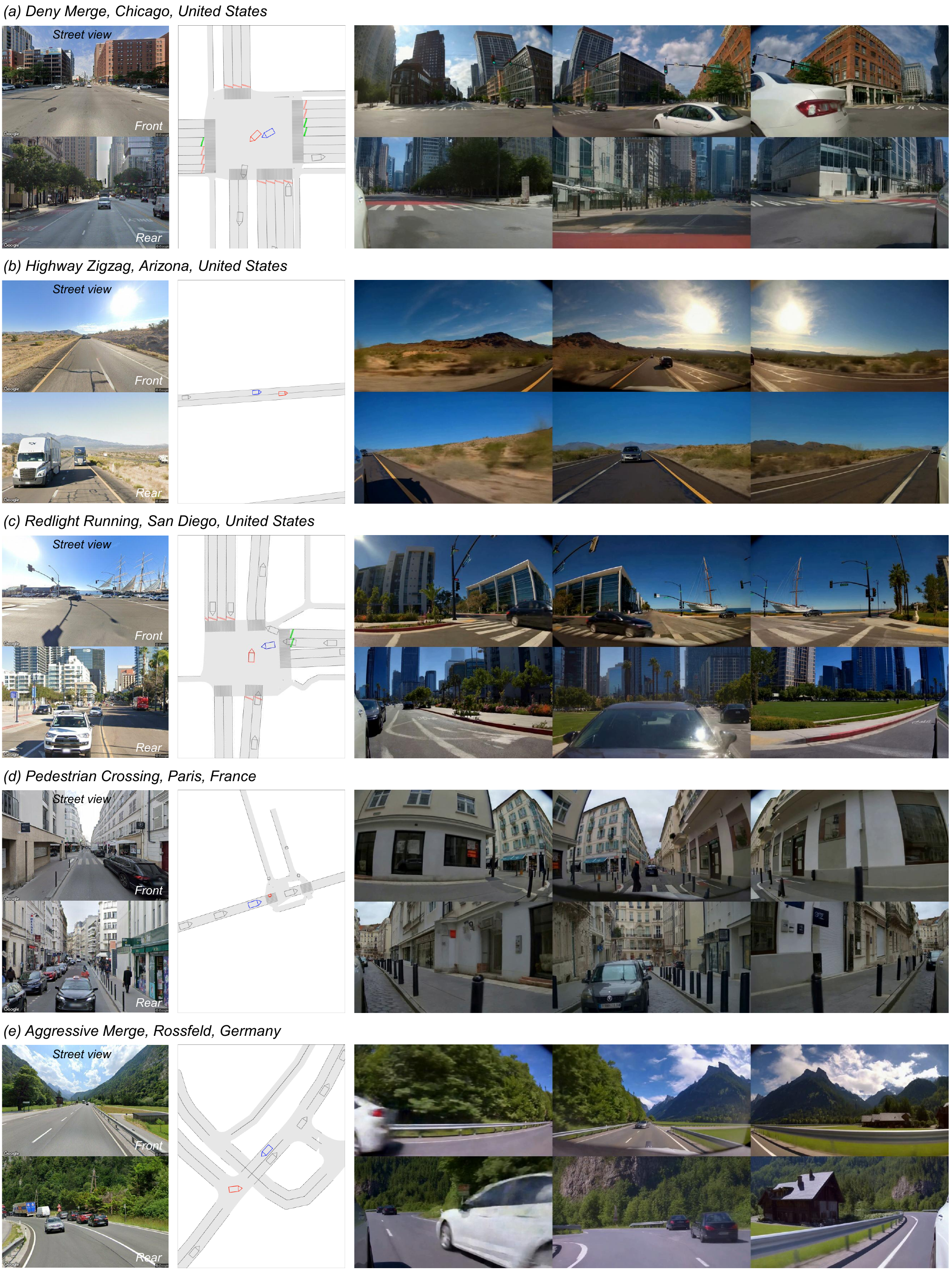}
    \vspace{-3mm}
	\caption{Examples of safety-critical data points generated worldwide. Scenario descriptions are listed in Table~\ref{tab:global_cases}. The first column shows the front and rear street-view images from Google Maps, and the second column shows trajectories (blue: ego, red: adversarial). The right panels present the snapshot of the generated multi-view videos, ordered from top-left to bottom-right as: front-left, front, front-right, rear-left, rear, and rear-right.}
	\label{Fig:global_case}
\end{figure*}

\subsection{Data Synthesis at Anywhere in the World}

Finally, to demonstrate the core capability of automatically generating safety-critical scenarios at arbitrary locations worldwide, we present five representative data points from different global coordinates, with descriptions provided in Table~\ref{tab:global_cases}. Fig.~\ref{Fig:global_case} illustrates the street-view images, simulated trajectories, and synthesized multi-view video snapshots. Note that TeraSim-World can generate a variety of safety-critical events, such as fail-to-yield, red-light running, or pedestrian conflicts, without manual intervention. Here we showcase only a subset of representative examples. 

It can be clearly observed that TeraSim-World generates adversarial interactions for traffic agents at each location, and synthesizes multi-view videos, where vehicle motions closely follow the simulated trajectories and visual styles highly align with the local environments. Together with the earlier case studies, these results span a wide range of road geometries (\emph{e.g.}, roundabouts, intersections, highways, urban streets, and narrow alleys), diverse traffic participants (including vulnerable road users such as pedestrians and cyclists), and varied global environments. These examples highlight the scalability of TeraSim-World in producing diverse, safety-critical driving data at anywhere in the world.

\begin{table*}[htb]
\centering
\caption{Worldwide Safety-Critical Scenarios Generated by TeraSim-World}
\vspace{-1em}
\label{tab:global_cases}
\begin{tabular}{p{0.16\linewidth} p{0.13\linewidth} p{0.64\linewidth}}
\toprule
\textbf{Coordinate} & \textbf{Location} & \textbf{Scenario Description} \\
\midrule
41.8967°N, 87.6325°W & Chicago, U.S. & A vehicle at the front-right blocks the ego vehicle’s attempt to merge during a left turn. \\
35.1610°N, 113.5990°W & Arizona, U.S. & A vehicle in the front drifts across lane markings in a zigzag manner on a highway. \\
32.7199°N, 117.1730°W & San Diego, U.S. & A vehicle from the left runs the red light and cuts across in front of the ego vehicle. \\
48.8594°N, 2.3061°E & Paris, France & A pedestrian suddenly crosses between two vehicles in front of the ego vehicle on a narrow alley. \\
47.5526°N, 12.7255°E & Rossfeld, Germany & A vehicle emerges from a right-hand side street and merges left across the lane on a suburban road. \\
\bottomrule
\end{tabular}
\vspace{-4mm}
\end{table*}






\section{Conclusions}
\label{Sec:6}

This paper presents TeraSim-World, an automated pipeline that unifies agent and sensor simulations to synthesize safety-critical data at anywhere in the world. It offers a scalable and cost-effective alternative to repeated data collection and extensive on-road testing for safety evaluation of E2E autonomous driving systems. 
Future work will focus on leveraging this pipeline to construct a large-scale synthesized dataset, and explore multi-modal sensor simulations and closed-loop AV testing for further applicability.





\section*{Acknowledgment}

The authors would like to thank Ruijie Chen, Yinghao Liu, Haoxiang Hong, and Miaomiao Dai for their assistance as interns in the development of certain modules.



\bibliographystyle{IEEEtran}
\bibliography{IEEEabrv,mybibfile}

\begin{thebibliography}{10}
\providecommand{\url}[1]{#1}
\csname url@samestyle\endcsname
\providecommand{\newblock}{\relax}
\providecommand{\bibinfo}[2]{#2}
\providecommand{\BIBentrySTDinterwordspacing}{\spaceskip=0pt\relax}
\providecommand{\BIBentryALTinterwordstretchfactor}{4}
\providecommand{\BIBentryALTinterwordspacing}{\spaceskip=\fontdimen2\font plus
\BIBentryALTinterwordstretchfactor\fontdimen3\font minus \fontdimen4\font\relax}
\providecommand{\BIBforeignlanguage}[2]{{%
\expandafter\ifx\csname l@#1\endcsname\relax
\typeout{** WARNING: IEEEtran.bst: No hyphenation pattern has been}%
\typeout{** loaded for the language `#1'. Using the pattern for}%
\typeout{** the default language instead.}%
\else
\language=\csname l@#1\endcsname
\fi
#2}}
\providecommand{\BIBdecl}{\relax}
\BIBdecl

\bibitem{liu2025autonomous}
H.~Liu, Z.~Cao, X.~Yan, S.~Feng, and Q.~Lu, ``Autonomous vehicles: A critical review (2004-2024) and a vision for the future,'' \emph{Authorea Preprints}, 2025.

\bibitem{chen2024end}
L.~Chen, P.~Wu, K.~Chitta, B.~Jaeger, A.~Geiger, and H.~Li, ``End-to-end autonomous driving: Challenges and frontiers,'' \emph{IEEE Transactions on Pattern Analysis and Machine Intelligence}, 2024.

\bibitem{sun2020scalability}
P.~Sun, H.~Kretzschmar, X.~Dotiwalla, A.~Chouard, V.~Patnaik, P.~Tsui, J.~Guo \emph{et~al.}, ``Scalability in perception for autonomous driving: Waymo open dataset,'' in \emph{Proceedings of the IEEE/CVF conference on computer vision and pattern recognition}, 2020, pp. 2446--2454.

\bibitem{caesar2020nuscenes}
H.~Caesar, V.~Bankiti, A.~H. Lang, S.~Vora, V.~E. Liong, Q.~Xu, A.~Krishnan, Y.~Pan, G.~Baldan, and O.~Beijbom, ``{nuScenes}: A multimodal dataset for autonomous driving,'' in \emph{Proceedings of the IEEE/CVF conference on computer vision and pattern recognition}, 2020, pp. 11\,621--11\,631.

\bibitem{treiber2013traffic}
M.~Treiber and A.~Kesting, \emph{Traffic flow dynamics}.\hskip 1em plus 0.5em minus 0.4em\relax Springer, 2013.

\bibitem{krajzewicz2012recent}
D.~Krajzewicz, J.~Erdmann, M.~Behrisch, L.~Bieker \emph{et~al.}, ``Recent development and applications of sumo-simulation of urban mobility,'' \emph{International journal on advances in systems and measurements}, vol.~5, no. 3\&4, pp. 128--138, 2012.

\bibitem{wu2024smart}
W.~Wu, X.~Feng \emph{et~al.}, ``{SMART}: Scalable multi-agent real-time motion generation via next-token prediction,'' \emph{Advances in Neural Information Processing Systems}, vol.~37, pp. 114\,048--114\,071, 2024.

\bibitem{yan2023learning}
X.~Yan, Z.~Zou, S.~Feng, H.~Zhu, H.~Sun, and H.~X. Liu, ``Learning naturalistic driving environment with statistical realism,'' \emph{Nature Communications}, vol.~14, no.~1, p. 2037, 2023.

\bibitem{zhang2025closed}
Z.~Zhang, P.~Karkus, M.~Igl, W.~Ding, Y.~Chen, B.~Ivanovic, and M.~Pavone, ``Closed-loop supervised fine-tuning of tokenized traffic models,'' in \emph{Proceedings of the Computer Vision and Pattern Recognition Conference}, 2025, pp. 5422--5432.

\bibitem{huang2024versatile}
Z.~Huang, Z.~Zhang, A.~Vaidya, Y.~Chen, C.~Lv, and J.~F. Fisac, ``Versatile scene-consistent traffic scenario generation as optimization with diffusion,'' \emph{arXiv preprint arXiv:2404.02524}, vol.~3, 2024.

\bibitem{zhong2023guided}
Z.~Zhong, D.~Rempe, D.~Xu, Y.~Chen, S.~Veer, T.~Che, B.~Ray, and M.~Pavone, ``Guided conditional diffusion for controllable traffic simulation,'' in \emph{2023 IEEE international conference on robotics and automation (ICRA)}.\hskip 1em plus 0.5em minus 0.4em\relax IEEE, 2023, pp. 3560--3566.

\bibitem{jiang2024scenediffuser}
M.~Jiang, Y.~Bai, A.~Cornman, C.~Davis, X.~Huang, H.~Jeon, S.~Kulshrestha, J.~Lambert \emph{et~al.}, ``Scenediffuser: Efficient and controllable driving simulation initialization and rollout,'' \emph{Advances in Neural Information Processing Systems}, vol.~37, pp. 55\,729--55\,760, 2024.

\bibitem{xu2025diffscene}
C.~Xu, A.~Petiushko, D.~Zhao, and B.~Li, ``Diffscene: Diffusion-based safety-critical scenario generation for autonomous vehicles,'' in \emph{Proceedings of the AAAI Conference on Artificial Intelligence}, vol.~39, no.~8, 2025, pp. 8797--8805.

\bibitem{chang2024safe}
W.-J. Chang, F.~Pittaluga, M.~Tomizuka, W.~Zhan, and M.~Chandraker, ``Safe-sim: Safety-critical closed-loop traffic simulation with diffusion-controllable adversaries,'' in \emph{European Conference on Computer Vision}.\hskip 1em plus 0.5em minus 0.4em\relax Springer, 2024, pp. 242--258.

\bibitem{wang2025rade}
J.~Wang, X.~Yan, Y.~Mu, H.~Sun, Z.~Cao, and H.~X. Liu, ``{RADE}: Learning risk-adjustable driving environment via multi-agent conditional diffusion,'' \emph{arXiv preprint arXiv:2505.03178}, 2025.

\bibitem{ho2022video}
J.~Ho, T.~Salimans, A.~Gritsenko, W.~Chan, M.~Norouzi, and D.~J. Fleet, ``Video diffusion models,'' \emph{Advances in neural information processing systems}, vol.~35, pp. 8633--8646, 2022.

\bibitem{wang2024drivedreamer}
X.~Wang, Z.~Zhu, G.~Huang, X.~Chen \emph{et~al.}, ``Drivedreamer: Towards real-world-drive world models for autonomous driving,'' in \emph{European conference on computer vision}.\hskip 1em plus 0.5em minus 0.4em\relax Springer, 2024, pp. 55--72.

\bibitem{gao2023magicdrive}
R.~Gao, K.~Chen, E.~Xie, L.~Hong, Z.~Li, D.-Y. Yeung, and Q.~Xu, ``Magicdrive: Street view generation with diverse 3d geometry control,'' \emph{arXiv preprint arXiv:2310.02601}, 2023.

\bibitem{zhao2025drivedreamer}
G.~Zhao, X.~Wang, Z.~Zhu, X.~Chen, G.~Huang, X.~Bao, and X.~Wang, ``Drivedreamer-2: Llm-enhanced world models for diverse driving video generation,'' in \emph{Proceedings of the AAAI Conference on Artificial Intelligence}, vol.~39, no.~10, 2025, pp. 10\,412--10\,420.

\bibitem{voleti2022mcvd}
V.~Voleti, A.~Jolicoeur-Martineau, and C.~Pal, ``Mcvd-masked conditional video diffusion for prediction, generation, and interpolation,'' \emph{Advances in neural information processing systems}, vol.~35, pp. 23\,371--23\,385, 2022.

\bibitem{hoppe2022diffusion}
T.~H{\"o}ppe, A.~Mehrjou, S.~Bauer, D.~Nielsen, and A.~Dittadi, ``Diffusion models for video prediction and infilling,'' \emph{arXiv preprint arXiv:2206.07696}, 2022.

\bibitem{hu2023gaia}
A.~Hu, L.~Russell, H.~Yeo, Z.~Murez, G.~Fedoseev, A.~Kendall, J.~Shotton, and G.~Corrado, ``Gaia-1: A generative world model for autonomous driving,'' \emph{arXiv preprint arXiv:2309.17080}, 2023.

\bibitem{wang2024driving}
Y.~Wang, J.~He, L.~Fan, H.~Li, Y.~Chen, and Z.~Zhang, ``Driving into the future: Multiview visual forecasting and planning with world model for autonomous driving,'' in \emph{Proceedings of the IEEE/CVF Conference on Computer Vision and Pattern Recognition}, 2024, pp. 14\,749--14\,759.

\bibitem{wen2024panacea}
Y.~Wen, Y.~Zhao, Y.~Liu, F.~Jia, Y.~Wang, C.~Luo, C.~Zhang, T.~Wang \emph{et~al.}, ``Panacea: Panoramic and controllable video generation for autonomous driving,'' in \emph{Proceedings of the IEEE/CVF Conference on Computer Vision and Pattern Recognition}, 2024, pp. 6902--6912.

\bibitem{yang2024drivearena}
X.~Yang, L.~Wen, Y.~Ma, J.~Mei, X.~Li, T.~Wei, W.~Lei, D.~Fu, P.~Cai \emph{et~al.}, ``Drivearena: A closed-loop generative simulation platform for autonomous driving,'' \emph{arXiv preprint arXiv:2408.00415}, 2024.

\bibitem{sun2025terasim}
H.~Sun, X.~Yan, Z.~Qiao, H.~Zhu, Y.~Sun \emph{et~al.}, ``{TeraSim}: Uncovering unknown unsafe events for autonomous vehicles through generative simulation,'' \emph{arXiv preprint arXiv:2503.03629}, 2025.

\bibitem{dosovitskiy2017carla}
A.~Dosovitskiy, G.~Ros, F.~Codevilla, A.~Lopez, and V.~Koltun, ``{CARLA}: An open urban driving simulator,'' in \emph{Conference on robot learning}.\hskip 1em plus 0.5em minus 0.4em\relax PMLR, 2017, pp. 1--16.

\bibitem{ren2025cosmos}
X.~Ren, Y.~Lu, T.~Cao, R.~Gao, S.~Huang, A.~Sabour, T.~Shen, T.~Pfaff, J.~Z. Wu, R.~Chen \emph{et~al.}, ``{Cosmos-Drive-Dreams}: Scalable synthetic driving data generation with world foundation models,'' \emph{arXiv preprint arXiv:2506.09042}, 2025.

\bibitem{agarwal2025cosmos}
N.~Agarwal, A.~Ali, M.~Bala, Y.~Balaji, E.~Barker, T.~Cai, P.~Chattopadhyay, Y.~Chen, Y.~Cui, Y.~Ding \emph{et~al.}, ``Cosmos world foundation model platform for physical {AI},'' \emph{arXiv preprint arXiv:2501.03575}, 2025.

\bibitem{montali2023waymo}
N.~Montali, J.~Lambert, P.~Mougin, A.~Kuefler, N.~Rhinehart, M.~Li, C.~Gulino, T.~Emrich, Z.~Yang, S.~Whiteson \emph{et~al.}, ``The waymo open sim agents challenge,'' \emph{Advances in Neural Information Processing Systems}, vol.~36, pp. 59\,151--59\,171, 2023.

\bibitem{feng2023dense}
S.~Feng, H.~Sun, X.~Yan, H.~Zhu, Z.~Zou, S.~Shen, and H.~X. Liu, ``Dense reinforcement learning for safety validation of autonomous vehicles,'' \emph{Nature}, vol. 615, no. 7953, pp. 620--627, 2023.

\bibitem{zhang2023adding}
L.~Zhang, A.~Rao, and M.~Agrawala, ``Adding conditional control to text-to-image diffusion models,'' in \emph{Proceedings of the IEEE/CVF international conference on computer vision}, 2023, pp. 3836--3847.

\end{thebibliography}

\end{document}